\pdfoutput=1
\documentclass[11pt,a4paper]{article}

\usepackage[hyperref]{acl2021}
\usepackage{times}
\usepackage{latexsym}

\usepackage{microtype}
\usepackage{float}
\usepackage{pgfplots}
\usepackage{makecell}

\aclfinalcopy

\title{UIT-ISE-NLP at SemEval-2021 Task 5: Toxic Spans Detection with BiLSTM-CRF and ToxicBERT Comment Classification}

\author{Son T. Luu \\
  University of Information Technology \\
  Vietnam National University \\
  Ho Chi Minh City, Vietnam \\
  \texttt{sonlt@uit.edu.vn} \\\And
  Ngan Nguyen \\
  University of Information Technology \\
  Vietnam National University \\
  Ho Chi Minh City, Vietnam \\
  \texttt{ngannlt@uit.edu.vn} \\}

\begin{document}
\maketitle

\begin{abstract}
We present our works on SemEval-2021 Task 5 about Toxic Spans Detection. This task aims to build a model for identifying toxic words in whole posts. We use the BiLSTM-CRF model combining with ToxicBERT Classification to train the detection model for identifying toxic words in posts. Our model achieves 62.23\% by F1-score on the Toxic Spans Detection task. 
\end{abstract}

\section{Introduction}
\label{intro}
Detecting toxic posts on social network sites is a crucial task for social media moderators in order to keep a clean and friendly space for online discussion. To identify whether a comment or post is toxic or not, social network administrators often read the whole comment or post. However, with a large number of lengthy posts, the administrators need assistance to locate toxic words in each post to decide whether a post is toxic or non-toxic instead of reading the whole post. The SemEval-2021 Task 5 \cite{pav2020semeval} provides a valuable dataset called Toxic Spans Detection dataset in order to train the model for detecting toxic words in lengthy posts. 

Based on the dataset from the shared task, we implement the machine learning model for detecting toxic words posts. Our model includes: the BiLSTM-CRF model \cite{lample-etal-2016-neural} for detecting the toxic spans in the post, and the ToxicBERT \cite{Detoxify} for classifying whether the post is toxic or not. Before training the model, we pre-process texts in posts and encode them by the GloVe word embedding \cite{pennington-etal-2014-glove}. Our model achieves 62.23\% on the test set provided by the task organizers. 

\section{Related works}
\label{related_works}
Many corpora are constructed for toxic speech detection problems. They consist of flat label and hierarchical label datasets. The flat label datasets only classify one label for each comment in the dataset (e.g., hate, offensive, clean), while hierarchical datasets can classify multiple aspects of the comment (e.g., hate about racism, hate about sexual oriented, hate about religion, and hate about disability). For flat label, we present several datasets including the two datasets which are provided by Waseem and Hovy \shortcite{waseem-hovy-2016-hateful} and Davidson et al. \shortcite{Davidson2017AutomatedHS} in English, the dataset of Albadi et al. \shortcite{8508247} in Arabic, and the dataset of by Alfina et al. \shortcite{8355039} in Indonesian. For the hierarchical label, we introduce the dataset constructed by Zampieri et al. \shortcite{zampieri-etal-2019-predicting} in English, the dataset provided by Fortuna et al. \shortcite{fortuna-etal-2019-hierarchically} in Portuguese, and the CONAN dataset by Chung et al. \shortcite{chung-etal-2019-conan}, which is the multilingual corpus (constructed in Italian, English, and French). 

In addition, many of shared tasks about hate speech and abusive languages are organized, such as the SemEval-2019 Task 5 (Multilingual) \cite{basile-etal-2019-semeval}, the SemEval-2019 Task 6 (English) \cite{zampieri-etal-2019-predicting}, the PolEval 2019 Shared task 6 (Polish) \cite{ptaszynski2019results}, the GermEval 2018 (Germany) \cite{wiegand2018overview}, EVALITA 2019 (Italian) \cite{bosco2018overview}, Toxic Comment Classification Challenge\footnote{https://www.kaggle.com/c/jigsaw-toxic-comment-classification-challenge}, and VLSP 2019 Shared task (Vietnamese) \cite{sonvx2019}. 

Besides, SOTA approaches like deep learning \cite{Badjatiya_2017} and transformers models \cite{isaksen-gamback-2020-using} are applied in the hate speech detection and toxic posts classification. However, these models only classify based on the whole posts or documents. For the Toxic Spans Detection task, we adapt the mechanism from Sequence tagging \cite{wang-etal-2020-contextualized} and Name entities Recognition \cite{lin-etal-2017-multi} for detecting toxic words from posts. 
\section{Dataset}
\label{dataset}
The dataset is provided from the SemEval-2021 Task 5: Toxic Spans Detection \cite{pav2020semeval}. It includes the training and the test sets. Both of them consist of two parts: the content of posts and the spans denoting the toxic words in the posts. Spans represent toxic words in the posts as a set of character indexes. Table \ref{tab_sample_data} illustrates several examples from the training set. 

\begin{table*}
    \centering
    \begin{tabular}{p{0.4cm}|p{8cm}|p{6cm}}
        \textbf{No} & \textbf{Posts} & \textbf{Spans} \\
        \hline
        1 & What a \textbf{knucklehead}. How can anyone not know this would be offensive?? & [7, 8, 9, 10, 11, 12, 13, 14, 15, 16, 17] \\
        \hline
        2 & I only use the word haole when \textbf{stupidity} and \textbf{arrogance} is involved and not all the time.  Excluding the POTUS of course. & [31, 32, 33, 34, 35, 36, 37, 38, 39, 45, 46, 47, 48, 49, 50, 51, 52, 53] \\
        \hline
        3 & \textbf{Such garbage logic by republicans} which will backfire and rush america into the great depression II & [0, 1, 2, 3, 4, 5, 6, 7, 8, 9, 10, 11, 12, 13, 14, 15, 16, 17, 18, 19, 20, 21, 22, 23, 24, 25, 26, 27, 28, 29, 30, 31, 32] \\
        \hline
        4 & what a \textbf{hypocrite} of bs,, tell us loser how you live without gasoline, plastic, medical needs and medications, all from OIL,, but you cant of course so you \textbf{ignorant} fools in your \textbf{hypocrisy} spew this bs & [7, 8, 9, 10, 11, 12, 13, 14, 15, 155, 156, 157, 158, 159, 160, 161, 162, 178, 179, 180, 181, 182, 183, 184, 185, 186] \\
        \hline
        5 & Exposing hypocrites like Trump and Pence is therapeutic for you? Good job! & [] \\
        \hline
    \end{tabular}
    \caption{Sample posts from the training set. The toxic span are highlighted as bold.}
    \label{tab_sample_data}
\end{table*}

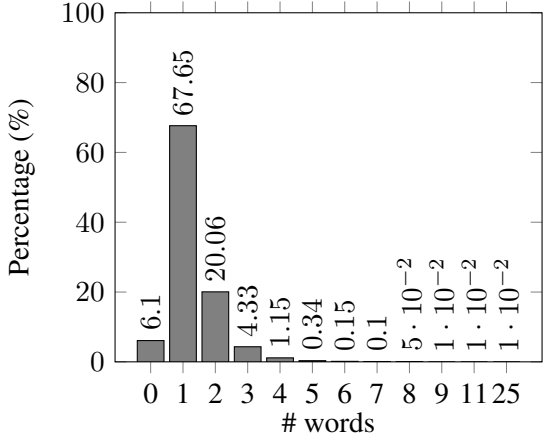
\begin{figure}[H]
    \centering
    \begin{tikzpicture}
        \begin{axis}[
                width=0.45\textwidth,
                ybar,
                enlarge y limits=0,
                ymax=100,
                symbolic x coords={0, 1, 2, 3, 4, 5, 6, 7, 8, 9, 11, 25},
                xtick=data,
                nodes near coords,
                ylabel={Percentage (\%)},
                xlabel={\# words},
                every node near coord/.append style={rotate=90, anchor=west},
            ]
            \addplot[ybar,fill=gray] coordinates {
                (0, 6.10)
                (1, 67.65)
                (2, 20.06)
                (3, 4.33)
                (4, 1.15)
                (5, 0.34)
                (6, 0.15)
                (7, 0.10)
                (8, 0.05)
                (9, 0.01)
                (11, 0.01)
                (25, 0.01)
            };
        \end{axis}
    \end{tikzpicture}
    \caption{Number of toxic words in spans for each post in the training set.}
    \label{fig_distrib_span_train}
\end{figure}

\begin{figure}[H]
    \centering
    \begin{tikzpicture}
        \begin{axis}[
                width=0.45\textwidth,
                ybar,
                enlarge y limits=0,
                ymax=100,
                symbolic x coords={0, 1, 2, 3, 4, 6, 7, 8, 9, 11, 25},
                xtick=data,
                nodes near coords,
                ylabel={Percentage (\%)},
                xlabel={\# words},
                every node near coord/.append style={rotate=90, anchor=west},
            ]
            \addplot[ybar,fill=gray] coordinates {
                (0, 19.70)
                (1, 70.35)
                (2, 8.60)
                (3, 0.08)
                (4, 0.04)
                (6, 0.01)
                (7, 0.005)
            };
        \end{axis}
    \end{tikzpicture}
    \caption{Number of toxic words in spans for each post in the test set.}
    \label{fig_distrib_span_test}
\end{figure}
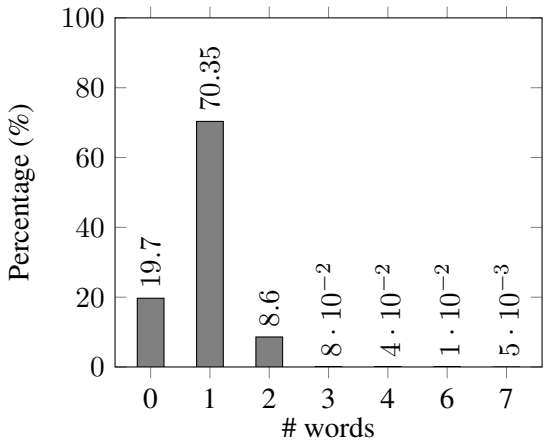

According to Table \ref{tab_sample_data}, a post contains multiple spans of toxic words. For each span, it contains a single word, a phrase, or a sentence. As described in Figure \ref{fig_distrib_span_train}, most of the spans in the training set are single words, which account for 67.65\%, while only 20.06\% of spans contains two words, and 6.1\% of spans is empty. Posts whose spans contain more than two words in the dataset are few. Especially in the training set, there is a post in which spans contain 25 words.

Besides, Figure \ref{fig_distrib_span_test} illustrates the number of toxic words in spans per post for the test set. Spans containing single words account for the highest percentage (70.35\%) in the test set, and are higher than in the training set, while the multiple-word spans are few. Also, the empty spans in the test set are higher than the training set, and the longest post in the test set contains only seven words. 
\section{System description}
\label{method}
\subsection{Data preparation}
With the given dataset from the SemEval-2021 Task 5 about Toxic Spans Detection \cite{pav2020semeval}, we firstly transform spans into a set of words. Then, we pre-process the posts as follows: (1) Segmenting the posts by the TweetTokenizer from nltk\footnote{\url{https://www.nltk.org/api/nltk.tokenize.html}}, and (2) Changing texts to lower case. 

\subsection{Feature extraction}
We use the glove.twitter.27b.25d word embedding\footnote{\url{https://nlp.stanford.edu/projects/glove/}} to construct the dictionary and encode the text of posts. Posts are encoded by the dictionary of the word embedding. The \textit{$<UNK>$} tokens are added if a word in posts is not found in the dictionary. To make sure all vectors are the same length, we add the \textit{$<PAD>$} token. Then, we set the maximum length of vectors equal to 128. Spans are transformed into a one-hot vector corresponding to each word in posts where toxic words are denoted as 1 and others are denoted as 0. Table \ref{tab_encode_data} illustrates an example of encoding data in our system.

\begin{table*}[ht]
    \begin{tabular}{p{1cm}|p{7.3cm}|p{6.4cm}}
        & \textbf{Original} & \textbf{Transformed} \\
        \hline
        \textbf{Text} & I only use the word haole when \textbf{stupidity} and \textbf{arrogance} is involved and not all the time. Excluding the POTUS of course. & \makecell*[{{p{6.4cm}}}]{['i', 'only', 'use', 'the', 'word', 'haole', ..] \\ {\textbf{Vector: }[12, 216, 718, 15, 894,..]} } \\
        \hline
        \textbf{Spans} & [31, 32, 33, 34, 35, 36, 37, 38, 39, 45, 46, 47, 48, 49, 50, 51, 52, 53] & \makecell*[{{p{6.4cm}}}]{{['i', 'only', 'use', 'the', 'word', 'haole', 'when', \textbf{'stupidity'}, 'and', \textbf{'arrogance'}, 'is', ...]} \\{\textbf{Vector: }[0, 0, 0, 0, 0, 0, 0, \textbf{1}, 0, \textbf{1}, 0, 0 ...]}} \\
        \hline
    \end{tabular}
    \caption{Example of encoding data into vectors.}
    \label{tab_encode_data}
\end{table*}

\subsection{Training models}

\begin{figure}[H]
    \centering
    \resizebox{.5\textwidth}{!}{
    \includegraphics{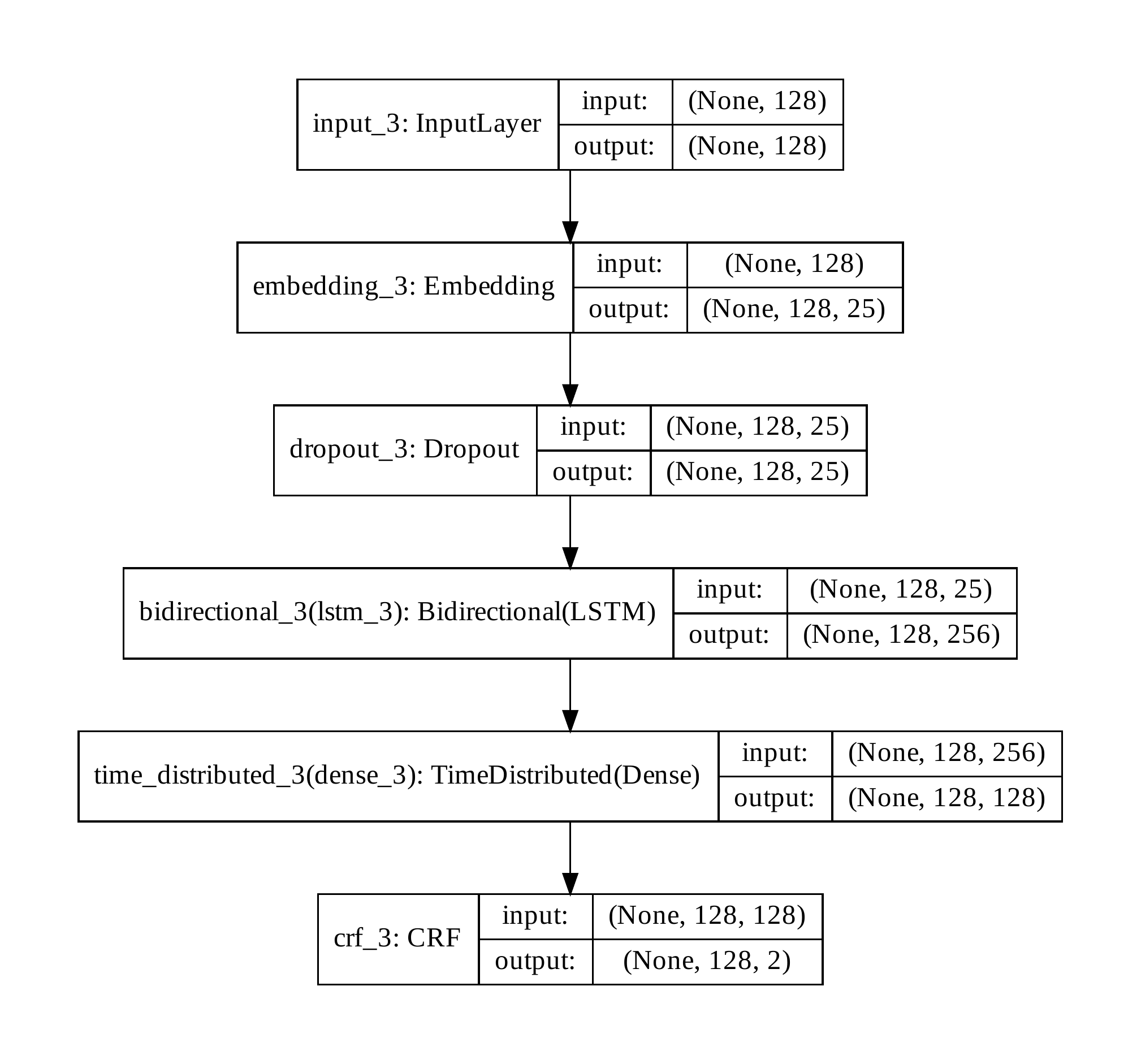}
    
    }
    \caption{The BiLSTM-CRF model architecture.}
    \label{fig_bilstmcrf_arch}
\end{figure}

\textbf{Detection model:} BiLSTM-CRF is a deep neural model used for Named-entity recognition task \cite{lample-etal-2016-neural}. We implement this model for the task of detecting toxic words in documents. The model includes three main layers: (1) The word representation layer uses embedding matrix from the GloVe word embedding, (2) The BiLSTM layer for sequence labeling, and (3) The Conditional Random Field (CRF) layer to control the probability of output labels. The output is a binary vector, in which each value determines whether a word is toxic or non-toxic. The architecture of BiLSTM-CRF is described in Figure \ref{fig_bilstmcrf_arch}.

\textbf{Classification model:} The ToxicBERT model (Detoxify) is introduced by Hanu and Unitary team \shortcite{Detoxify} with the purpose to stop online abusive comments. It is a pre-trained model and is easy to use by using transformers library\footnote{\url{https://huggingface.co/unitary/toxic-bert}}. The model is trained on the Toxic Comments Classification Challenge datasets provided by Jigsaw. 

\begin{figure}[H]
    \centering
    \includegraphics[height=9cm,width=8cm]{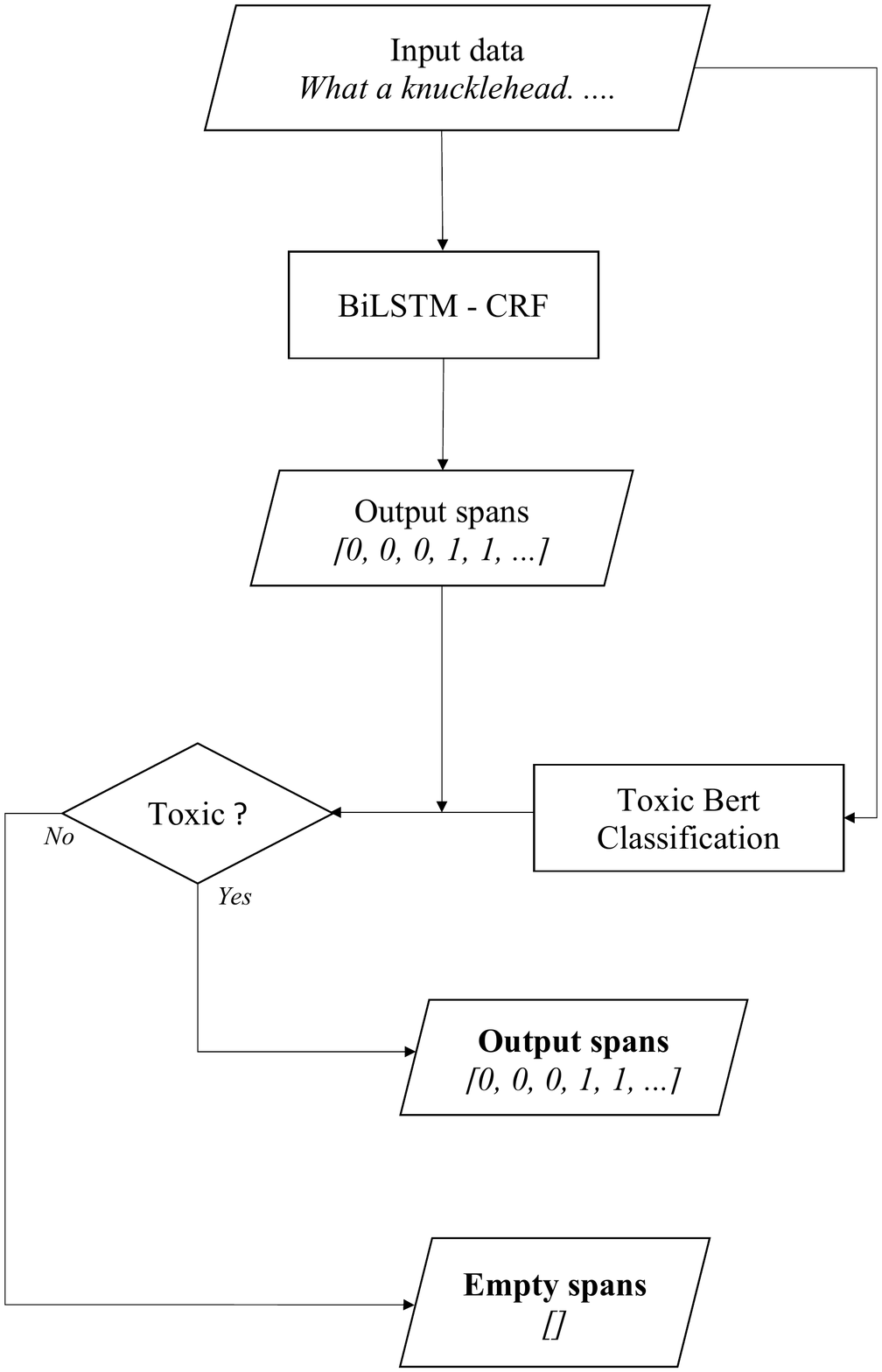}
    \caption{Our system architecture.}
    \label{fig_system_arch}
\end{figure}

Our system combines the detection and classification model together. The detection model (BiLSTM-CRF) returns the toxic spans from the post, while the classification model (ToxicBERT) classifies whether a post is toxic or non-toxic. If a post is non-toxic, the classification model returns an empty span. By contrast, it reserves the spans of the detection model. Then, predicted spans are decoded to character indexes for submission. Our system is illustrated in Figure \ref{fig_system_arch}

\section{Experimental results}
\label{results}
\subsection{Evaluation metric}
The variant version of F1-score is used to evaluate the results of the competition \cite{da-san-martino-etal-2019-fine}. Let T is the total of post in the dataset, $T=[t_1, t_2, ..., t_n]$, $n$ is the number of posts, $A$ is spans given by the model, and $G$ is ground truth spans. 

The F1-score over the dataset is defined as:
\begin{equation}
    \label{eq_f1_score}
    \frac{1}{|T|}\sum_{t}^{T} F_1^{t} = 2*\frac{P^t(A, G)*R^t(A,G)}{P^t(A, G)+R^t(A,G)}
\end{equation}

In the Equation \ref{eq_f1_score}, $P^t$ determines the precision, and $R^t$ determines the recall of the post t. The precision and recall are calculated as Equation \ref{eq_precision_score} and Equation \ref{eq_recall_score}, respectively. The $S^t$ in both Equation \ref{eq_precision_score} and Equation \ref{eq_recall_score} is set of toxic characters of post t (span). 

\begin{equation}
    \label{eq_precision_score}
    P^t(A,G) = \frac{|S^t_A \cap S^t_G|}{S^t_A}
\end{equation}

\begin{equation}
    \label{eq_recall_score}
    R^t(A,G) = \frac{|S^t_A \cap S^t_G|}{S^t_G}
\end{equation}

\subsection{Main result from task}
\begin{table}[H]
    \centering
    \begin{tabular}{|c|c|}
        \hline
        \textbf{Model} & \textbf{Result (\%)} \\
        \hline
        BiLSTM-CRF & 61.32 \\
        \hline
        \textbf{BiLSTM-CRF + ToxicBERT} & \textbf{62.23} \\
        \hline
        Top 1 rank & \textbf{70.83} \\
        \hline
    \end{tabular}
    \caption{Experimental results obtained by our system in comparison with Top 1 rank result.}
    \label{tab_result_final}
\end{table}

According to Table \ref{tab_result_final}, when only BiLSTM-CRF is used, the result by F1 score is 61.32\%. The result increases up to 62.23\% when we applied ToxicBERT Classifier, and this is our final result of the shared task (ranked $63^{th}$ among 92 teams). 

\subsection{Error analysis}

\begin{table*}[ht!]
    \centering
    \begin{tabular}{p{0.5cm}|p{3.5cm}|p{3.2cm}|p{3.2cm}|p{3.2cm}}
        \textbf{No. } & \textbf{Posts} & \textbf{Gold span} & \textbf{\makecell{Prediction span: \\ BiLSTM-CRF}} & \textbf{\makecell*[{{p{3cm}}}]{Prediction span: \\ BiLSTM-CRF+ ToxicBERT}}\\
        \hline
        1 & Indeed, people the world over, all know that President Trump is a loser! & [] & [66, 67, 68, 69, 70] & [] \\
        \hline
        2 & Idiot "miner" in the photo isn't even wearing a dust mask. MSHA needs to inspect this operation for more worker safety violations. & [0, 1, 2, 3, 4] & [-1, 0, 1, 2, 3] & [-1, 0, 1, 2, 3] \\
        \hline
        3 & People who are anti-immigration are weak, lying, racist whiners. & [49, 50, 51, 52, 53, 54, 55, 56, 57, 58, 59, 60, 61, 62] & [] & [] \\
        \hline
        4 & Uh-No, keep voting for failed Liberal idiocy that guarantees results ala Detroit, Chicago, etc.  You'll wish your body had only some crap rather than gangbanger gunfire. & [38, 39, 40, 41, 42, 43] & [38, 39, 40, 41, 42, 43, 133, 134, 135, 136] & [38, 39, 40, 41, 42, 43, 133, 134, 135, 136] \\
        \hline
        5 & What is he going to do about those toxic mercury florescent bulbs Bush and Gore pushed on the stupid American public? & [94, 95, 96, 97, 98, 99, 100, 101, 102, 103, 104, 105, 106, 107, 108, 109, 110, 111, 112, 113, 114, 115] & [94, 95, 96, 97, 98, 99] & [94, 95, 96, 97, 98, 99] \\
        \hline
    \end{tabular}
    \caption{Several wrong predictions on the test set by our system.}
    \label{tab_error_analysis}
\end{table*}

According to Table \ref{tab_error_analysis}, the appearance of the ToxicBERT classifier can make a better prediction for the non-toxic posts (See example No. 1). This increases the performance of our system, however, not significantly, because the number of empty toxic span comments in the test set is not too much (as described in Section \ref{dataset}). Apart from empty spans, wrong predictions are fell into the case as example No.3. Although the ToxicBERT model predicts this example as non-toxic, the BiLSTM-CRF model predicts it as empty spans, and thus the result is empty spans according to the result of the BiLSTM-CRF model. For example No.2, the system returns wrong spans for the first span in the post. Finally, our system cannot predict well for spans that contain more than two words. It returns spans, but not enough, as shown in examples No.4 and No.5 from Table \ref{tab_error_analysis}. 
\section{Conclusion}
\label{conclusion}
We use the BiLSTM-CRF and ToxicBERT models for detecting toxic words in the posts. Our model achieves 62.23\% by F1-score from the competition. From the error analysis, we found that our model predicts well just for single-word spans and empty spans.

In further researches, we improve the performance of the detection model by applying the attention mechanism and using the character-level representation combining with word-level representation. Character-level models like CharBERT \cite{ma-etal-2020-charbert} is a potential approach to increase the performance of toxic spans detection tasks.

\bibliographystyle{acl_natbib}
\bibliography{acl2021}

\begin{thebibliography}{22}
\expandafter\ifx\csname natexlab\endcsname\relax\def\natexlab#1{#1}\fi

\bibitem[{{Albadi} et~al.(2018){Albadi}, {Kurdi}, and {Mishra}}]{8508247}
N.~{Albadi}, M.~{Kurdi}, and S.~{Mishra}. 2018.
\newblock Are they our brothers? analysis and detection of religious hate
  speech in the arabic twittersphere.
\newblock In \emph{IEEE/ACM International Conference on Advances in Social
  Networks Analysis and Mining (ASONAM)}, pages 69--76.

\bibitem[{{Alfina} et~al.(2017){Alfina}, {Mulia}, {Fanany}, and
  {Ekanata}}]{8355039}
I.~{Alfina}, R.~{Mulia}, M.~I. {Fanany}, and Y.~{Ekanata}. 2017.
\newblock Hate speech detection in the indonesian language: A dataset and
  preliminary study.
\newblock In \emph{2017 International Conference on Advanced Computer Science
  and Information Systems (ICACSIS)}, pages 233--238.

\bibitem[{Badjatiya et~al.(2017)Badjatiya, Gupta, Gupta, and
  Varma}]{Badjatiya_2017}
Pinkesh Badjatiya, Shashank Gupta, Manish Gupta, and Vasudeva Varma. 2017.
\newblock Deep learning for hate speech detection in tweets.
\newblock \emph{Proceedings of the 26th International Conference on World Wide
  Web Companion - WWW ’17 Companion}.

\bibitem[{Basile et~al.(2019)Basile, Bosco, Fersini, Nozza, Patti,
  Rangel~Pardo, Rosso, and Sanguinetti}]{basile-etal-2019-semeval}
Valerio Basile, Cristina Bosco, Elisabetta Fersini, Debora Nozza, Viviana
  Patti, Francisco~Manuel Rangel~Pardo, Paolo Rosso, and Manuela Sanguinetti.
  2019.
\newblock \href {https://doi.org/10.18653/v1/S19-2007} {{S}em{E}val-2019 task
  5: Multilingual detection of hate speech against immigrants and women in
  {T}witter}.
\newblock In \emph{Proceedings of the 13th International Workshop on Semantic
  Evaluation}, pages 54--63, Minneapolis, Minnesota, USA. Association for
  Computational Linguistics.

\bibitem[{Bosco et~al.(2018)Bosco, Felice, Poletto, Sanguinetti, and
  Maurizio}]{bosco2018overview}
Cristina Bosco, Dell'Orletta Felice, Fabio Poletto, Manuela Sanguinetti, and
  Tesconi Maurizio. 2018.
\newblock Overview of the evalita 2018 hate speech detection task.
\newblock In \emph{EVALITA 2018-Sixth Evaluation Campaign of Natural Language
  Processing and Speech Tools for Italian}, volume 2263, pages 1--9. CEUR.

\bibitem[{Chung et~al.(2019)Chung, Kuzmenko, Tekiroglu, and
  Guerini}]{chung-etal-2019-conan}
Yi-Ling Chung, Elizaveta Kuzmenko, Serra~Sinem Tekiroglu, and Marco Guerini.
  2019.
\newblock \href {https://doi.org/10.18653/v1/P19-1271} {{CONAN} - {CO}unter
  {NA}rratives through nichesourcing: a multilingual dataset of responses to
  fight online hate speech}.
\newblock In \emph{Proceedings of the 57th Annual Meeting of the Association
  for Computational Linguistics}, pages 2819--2829, Florence, Italy.
  Association for Computational Linguistics.

\bibitem[{Da~San~Martino et~al.(2019)Da~San~Martino, Yu, Barr{\'o}n-Cede{\~n}o,
  Petrov, and Nakov}]{da-san-martino-etal-2019-fine}
Giovanni Da~San~Martino, Seunghak Yu, Alberto Barr{\'o}n-Cede{\~n}o, Rostislav
  Petrov, and Preslav Nakov. 2019.
\newblock \href {https://doi.org/10.18653/v1/D19-1565} {Fine-grained analysis
  of propaganda in news article}.
\newblock In \emph{Proceedings of the 2019 Conference on Empirical Methods in
  Natural Language Processing and the 9th International Joint Conference on
  Natural Language Processing (EMNLP-IJCNLP)}, pages 5636--5646, Hong Kong,
  China. Association for Computational Linguistics.

\bibitem[{Davidson et~al.(2017)Davidson, Warmsley, Macy, and
  Weber}]{Davidson2017AutomatedHS}
Thomas Davidson, Dana Warmsley, Michael Macy, and Ingmar Weber. 2017.
\newblock Automated hate speech detection and the problem of offensive
  language.

\bibitem[{Fortuna et~al.(2019)Fortuna, Rocha~da Silva, Soler-Company, Wanner,
  and Nunes}]{fortuna-etal-2019-hierarchically}
Paula Fortuna, Jo{\~a}o Rocha~da Silva, Juan Soler-Company, Leo Wanner, and
  S{\'e}rgio Nunes. 2019.
\newblock A hierarchically-labeled {P}ortuguese hate speech dataset.
\newblock In \emph{Proceedings of the Third Workshop on Abusive Language
  Online}, pages 94--104, Florence, Italy. Association for Computational
  Linguistics.

\bibitem[{Hanu and {Unitary team}(2020)}]{Detoxify}
Laura Hanu and {Unitary team}. 2020.
\newblock Detoxify.
\newblock Github. https://github.com/unitaryai/detoxify.

\bibitem[{Isaksen and Gamb{\"a}ck(2020)}]{isaksen-gamback-2020-using}
Vebj{\o}rn Isaksen and Bj{\"o}rn Gamb{\"a}ck. 2020.
\newblock Using transfer-based language models to detect hateful and offensive
  language online.
\newblock In \emph{Proceedings of the Fourth Workshop on Online Abuse and
  Harms}, Online. Association for Computational Linguistics.

\bibitem[{Lample et~al.(2016)Lample, Ballesteros, Subramanian, Kawakami, and
  Dyer}]{lample-etal-2016-neural}
Guillaume Lample, Miguel Ballesteros, Sandeep Subramanian, Kazuya Kawakami, and
  Chris Dyer. 2016.
\newblock \href {https://doi.org/10.18653/v1/N16-1030} {Neural architectures
  for named entity recognition}.
\newblock In \emph{Proceedings of the 2016 Conference of the North {A}merican
  Chapter of the Association for Computational Linguistics: Human Language
  Technologies}, pages 260--270, San Diego, California. Association for
  Computational Linguistics.

\bibitem[{Lin et~al.(2017)Lin, Xu, Luo, and Zhu}]{lin-etal-2017-multi}
Bill~Y. Lin, Frank Xu, Zhiyi Luo, and Kenny Zhu. 2017.
\newblock \href {https://doi.org/10.18653/v1/W17-4421} {Multi-channel
  {B}i{LSTM}-{CRF} model for emerging named entity recognition in social
  media}.
\newblock In \emph{Proceedings of the 3rd Workshop on Noisy User-generated
  Text}, pages 160--165, Copenhagen, Denmark. Association for Computational
  Linguistics.

\bibitem[{Ma et~al.(2020)Ma, Cui, Si, Liu, Wang, and
  Hu}]{ma-etal-2020-charbert}
Wentao Ma, Yiming Cui, Chenglei Si, Ting Liu, Shijin Wang, and Guoping Hu.
  2020.
\newblock \href {https://doi.org/10.18653/v1/2020.coling-main.4} {{C}har{BERT}:
  Character-aware pre-trained language model}.
\newblock In \emph{Proceedings of the 28th International Conference on
  Computational Linguistics}, pages 39--50, Barcelona, Spain (Online).
  International Committee on Computational Linguistics.

\bibitem[{Pavlopoulos et~al.(2021)Pavlopoulos, Laugier, Sorensen, and
  Androutsopoulos}]{pav2020semeval}
John Pavlopoulos, Léo Laugier, Jeffrey Sorensen, and Ion Androutsopoulos.
  2021.
\newblock Semeval-2021 task 5: Toxic spans detection (to appear).
\newblock In \emph{Proceedings of the 15th International Workshop on Semantic
  Evaluation}.

\bibitem[{Pennington et~al.(2014)Pennington, Socher, and
  Manning}]{pennington-etal-2014-glove}
Jeffrey Pennington, Richard Socher, and Christopher Manning. 2014.
\newblock \href {https://doi.org/10.3115/v1/D14-1162} {{G}lo{V}e: Global
  vectors for word representation}.
\newblock In \emph{Proceedings of the 2014 Conference on Empirical Methods in
  Natural Language Processing ({EMNLP})}, pages 1532--1543, Doha, Qatar.
  Association for Computational Linguistics.

\bibitem[{Ptaszynski et~al.(2019)Ptaszynski, Pieciukiewicz, and
  Dyba{\l}a}]{ptaszynski2019results}
Michal Ptaszynski, Agata Pieciukiewicz, and Pawe{\l} Dyba{\l}a. 2019.
\newblock Results of the poleval 2019 shared task 6: First dataset and open
  shared task for automatic cyberbullying detection in polish twitter.

\bibitem[{Vu et~al.(2019)Vu, Vu, Tran, Le-Cong, and Nguyen}]{sonvx2019}
Xuan-Son Vu, Thanh Vu, Mai-Vu Tran, Thanh Le-Cong, and Huyen T~M. Nguyen. 2019.
\newblock {HSD} shared task in {VLSP} campaign 2019: Hate speech detection for
  social good.
\newblock In \emph{Proceedings of VLSP 2019}.

\bibitem[{Wang et~al.(2020)Wang, Zhang, Ma, Wang, and
  Xiao}]{wang-etal-2020-contextualized}
Yan Wang, Jiayu Zhang, Jun Ma, Shaojun Wang, and Jing Xiao. 2020.
\newblock \href {https://www.aclweb.org/anthology/2020.sigdial-1.23}
  {Contextualized emotion recognition in conversation as sequence tagging}.
\newblock In \emph{Proceedings of the 21th Annual Meeting of the Special
  Interest Group on Discourse and Dialogue}, pages 186--195, 1st virtual
  meeting. Association for Computational Linguistics.

\bibitem[{Waseem and Hovy(2016)}]{waseem-hovy-2016-hateful}
Zeerak Waseem and Dirk Hovy. 2016.
\newblock Hateful symbols or hateful people? predictive features for hate
  speech detection on {T}witter.
\newblock In \emph{Proceedings of the {NAACL} Student Research Workshop}, pages
  88--93, San Diego, California. Association for Computational Linguistics.

\bibitem[{Wiegand et~al.(2018)Wiegand, Siegel, and
  Ruppenhofer}]{wiegand2018overview}
Michael Wiegand, Melanie Siegel, and Josef Ruppenhofer. 2018.
\newblock Overview of the germeval 2018 shared task on the identification of
  offensive language.

\bibitem[{Zampieri et~al.(2019)Zampieri, Malmasi, Nakov, Rosenthal, Farra, and
  Kumar}]{zampieri-etal-2019-predicting}
Marcos Zampieri, Shervin Malmasi, Preslav Nakov, Sara Rosenthal, Noura Farra,
  and Ritesh Kumar. 2019.
\newblock Predicting the type and target of offensive posts in social media.
\newblock In \emph{Proceedings of the 2019 Conference of the North {A}merican
  Chapter of the Association for Computational Linguistics: Human Language
  Technologies, Volume 1 (Long and Short Papers)}, Minneapolis, Minnesota.

\end{thebibliography}

\end{document}